\documentclass[10pt,twocolumn,letterpaper]{article}

\usepackage{wacv}
\usepackage{times}
\usepackage{epsfig}
\usepackage{graphicx}
\usepackage{amsmath}
\usepackage{amssymb}
\usepackage{helvet}
\usepackage{courier}
\usepackage{amsmath}
\usepackage{amssymb}
\usepackage{graphicx}
\usepackage{subfigure}
\usepackage{longtable}
\usepackage{rotating}
\usepackage{multirow}
\usepackage{multicol}
\usepackage{xcolor}
\usepackage{fancyhdr}
\usepackage{enumerate}
\usepackage{float}
\usepackage{balance}
\usepackage{url}
\usepackage{booktabs}

\def\wacvPaperID{125} 

\wacvfinalcopy 

\ifwacvfinal
\def\assignedStartPage{1} 
\fi

\ifwacvfinal
\usepackage[breaklinks=true,bookmarks=false]{hyperref}
\else
\usepackage[pagebackref=true,breaklinks=true,colorlinks,bookmarks=false]{hyperref}
\fi

\ifwacvfinal
\else
\fi

\begin{document}

\title{Adversarial Dual Distinct Classifiers for Unsupervised Domain Adaptation}


\author{Taotao Jing$^1$, ~Zhengming Ding$^2$\\
$^1$Department of ECE, Indiana University-Purdue University Indianapolis, USA \\$^2$Department of CIT, Indiana University-Purdue University Indianapolis, USA\\
{jingt, zd2}@iu.edu}



\maketitle
\begin{abstract}
   Unsupervised Domain adaptation (UDA) attempts to recognize the unlabeled target samples by building a learning model from a differently-distributed labeled source domain. Conventional UDA concentrates on extracting domain-invariant features through deep adversarial networks. However, most of them seek to match the different domain feature distributions, without considering the task-specific decision boundaries across various classes. In this paper, we propose a novel Adversarial Dual Distinct Classifiers Network (AD$^2$CN) to align the source and target domain data distribution simultaneously with matching task-specific category boundaries. To be specific, a domain-invariant feature generator is exploited to embed the source and target data into a latent common space with the guidance of discriminative cross-domain alignment. Moreover, we naturally design two different structure classifiers to identify the unlabeled target samples over the supervision of the labeled source domain data. Such dual distinct classifiers with various architectures can capture diverse knowledge of the target data structure from different perspectives. Extensive experimental results on several cross-domain visual benchmarks prove the model's effectiveness by comparing it with other state-of-the-art UDA.
\end{abstract}

\section{Introduction}

Deep neural networks (DNNs) have made significant progress with the help of numerous well-labeled training data and achieved remarkable performance improvement on various tasks \cite{krizhevsky2012imagenet,simonyan2014very}. However, massive amounts of annotated training data are not always available due to the dramatically expensive data collecting and annotating costs. Domain adaptation (DA) has attracted extremely increasing attention because it focuses on a frequent and real-world issue when we have no access to massive labeled target domain training data \cite{long2013transfer,li2019joint,morerio2020generative}. The mechanism of domain adaptation is to uncover the common latent factors across the source and target domains and reduce both the marginal and conditional mismatch in terms of the feature space between domains. Following this, different domain adaptation techniques have been developed, including feature alignment and classifier adaptation \cite{tsai2016domain,hou2016unsupervised,yan2017mind}.

Recent research efforts on domain adaptation have already shown promising performance via seeking an effective domain-invariant feature extractor across two domains so that the source knowledge could be adapted to facilitate the recognition task in the target domain \cite{herath2017learning,venkateswara2017deep,pizzati2020domain,iqbal2020mlsl,hsu2020progressive}. The idea is to deploy cross-domain matching losses to guide the domain-invariant feature learning. First of all, the discrepancy loss (e.g., maximum mean discrepancy (MMD)) is one of the most widely-used strategies to measure the distribution difference across the source and target domains \cite{borgwardt2006integrating,gretton2007kernel}. Along this, many DA approaches explore to design a class-wise MMD by incorporating the pseudo labels of target data \cite{hou2016unsupervised,morerio2020generative}. Secondly, the adversarial loss has been successfully applied to eliminate the domain shifts on the feature or pixel level \cite{ganin2014unsupervised,saito2018maximum,lee2019sliced,zhang2019domain}, where one domain discriminator or more are trained with a feature generator in an adversarial manner. Moreover, various reconstruction penalties are proposed on target samples to obtain the target specific structures, e.g., iCAN \cite{zhang2018collaborative}. However, most existing domain adaptation methods suffer from explicitly matching source and target domains distribution by only considering the domain-wise adaptation while ignoring the alignment of task-specific category boundaries.


\begin{figure}[t]
    \centering
    \hspace{-2mm}\includegraphics[width=0.49\textwidth]{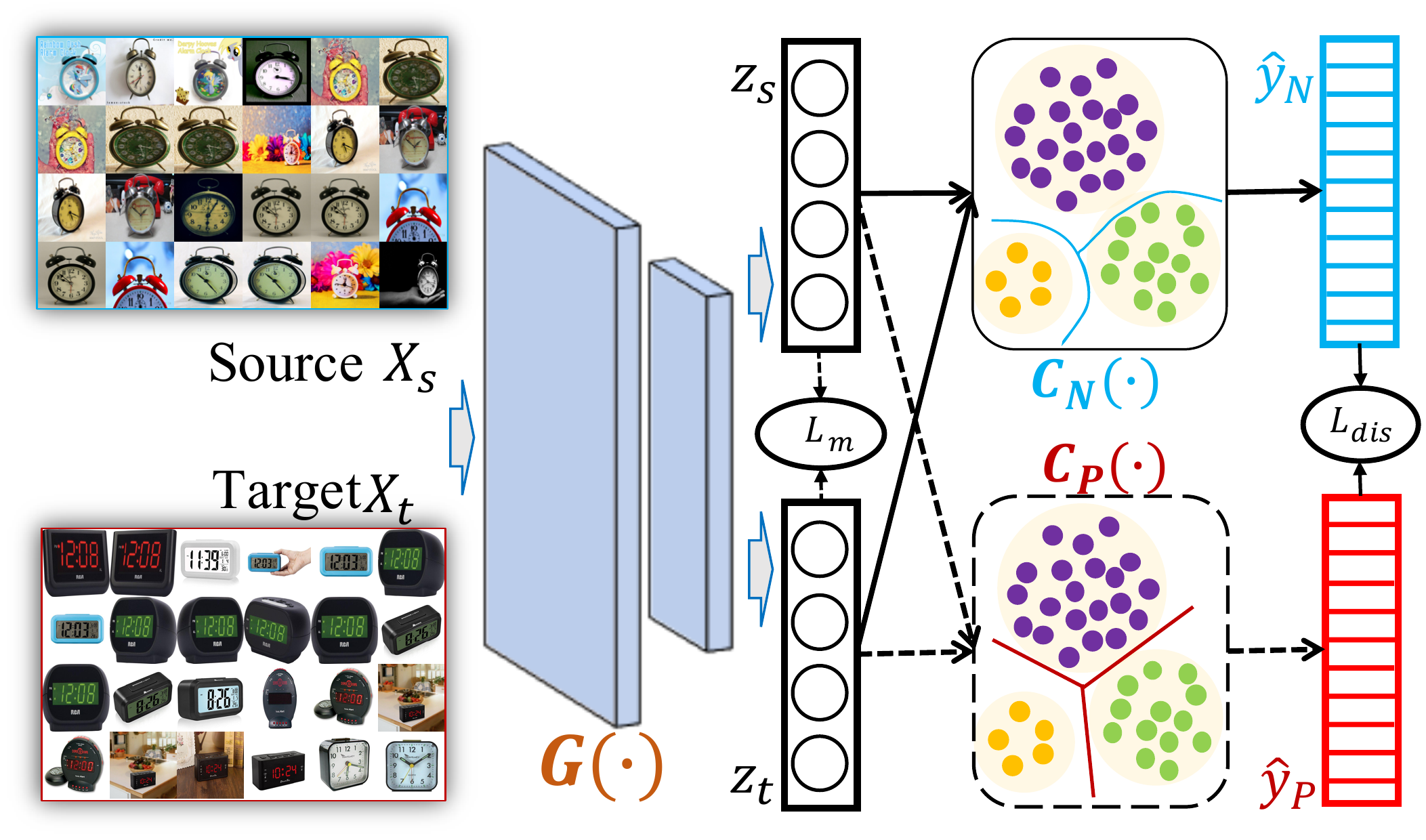}\vspace{-1mm}
    \caption{Framework overview of our proposed model, where $G(\cdot)$ is the domain-invariant embedding features generator, $C_N(\cdot)$ denotes the fully-connected neural networks classifier (solid line) and $C_P(\cdot)$ means the prototypical classifier (dash line). $\mathcal{L}_m$ and $\mathcal{L}_{dis}$ are explored to align the feature and prediction distribution differences across two domains and dual classifiers, respectively.}
    \label{fig:framework}\vspace{-3mm}
\end{figure}

To address this issue, some recent DA works aim to consider the task-specific category-level alignment jointly \cite{saito2018maximum,lee2019sliced,li2019joint}. Along this line, Saito \etal present Maximum Classifier Discrepancy (MCD) with two task-specific classifiers to detect category boundaries and jointly align features distribution and category boundaries across domains \cite{saito2018maximum}. Following this, Lee \etal propose Sliced Wasserstein Discrepancy (SWD) as a new probability distribution discrepancy measurement to capture the natural notion of dissimilarity between the outputs of task-specific classifiers \cite{lee2019sliced}. Later on, \cite{zhang2019domain} promotes Domain-Symmetric Networks (SymNets) as well as a two-level (feature-level and category-level) domain confusion scheme to drive the learning of intermediate features to be invariant at the corresponding categories of the two domains. These method benefit from various strategies to maximize the disparity of the dual classifiers prediction results, however, considering the utterly same architecture classifiers not only limits the features distribution knowledge obtained from different perspectives, but also suffers from the risk that the two task-specific classifiers may result in the similar class-wise boundaries, especially when the imbalanced data distribution across various categories.  

In this paper, we propose a novel Adversarial Dual Distinct Classifiers Network (AD$^2$CN) with two different-architecture classifiers, e.g., Neural Networks Classifier and Prototypical Classifier, to facilitate the alignment of both domain distributions and category decision boundaries (Fig. \ref{fig:framework}). To our best knowledge, it is a pioneering work to explore dual different structure classifiers in domain adaptation. The general idea is to explore adversarial training over two different architecture classifiers on the output of one domain-invariant feature generator. To sum up, we highlight the three-fold contributions of this paper as follows:
\begin{itemize}
	\item We exploit dual different architecture task-specific classifiers over source supervision to exploit the task-specific decision boundaries on the target domain. With different properties of dual classifiers in prediction, we have a better chance of capturing ground-truth classifier decision boundaries for the target domain.
    \item We propose a novel discriminative cross-domain alignment loss and \textit{Importance Guided Optimization} strategy to mitigate the cross-domain mismatching. This will facilitate the process of aligning the domain-invariant embedding features distribution across domains, and eliminate the distraction of misestimated target samples at the beginning of optimizing.
    \item We adopt a discrepancy loss to maximally improve the prediction performance of dual classifiers in coupling the cross-domain label distributions, which is trained in an adversarial way with domain-invariant feature generator and dual classifiers. Thus, they can benefit from each other to boost the target learning task.
\end{itemize}

\section{Related Work}

Domain adaptation (DA) has been extensively studied recently, which casts a light when there are no or limited labels in target domain and shows very promising performance in different vision applications \cite{zhang2019domain,li2019joint,long2016unsupervised,xu2019larger,lee2019sliced}.

With the renaissance of deep neural networks, deep DA methods successfully embed DA into deep learning pipelines by either minimizing an appropriate distribution distance metric \cite{long2018conditional} or leveraging adversarial technologies to generate domain-invariant representations \cite{saito2018maximum,chen2019transferability}. The idea behind this is to incorporate domain alignment strategies at the top layers to explicitly solve the enlarged domain discrepancy resulted from traditional deep learning models. To name a few, Long \etal proposed Domain Adaptation Network (DAN) to incorporate multiple kernel MMD distances across domains among the last three task-specific layers \cite{long2015learning}. Long \etal presented a joint adaptation network (JAN) as well as a joint MMD criterion \cite{long2017deep}. Another strategy is to leverage generative adversarial networks (GAN) \cite{goodfellow2014generative} to couple the cross-domain discrepancy in an adversarial manner \cite{ganin2014unsupervised,saito2018maximum,zhang2018collaborative,zhang2019domain}. Such techniques aim to train a domain discriminator to 
differentiate source and target samples, while the feature generator will deceive the domain discriminator, such that the domain-invariant features will be produced. Ganin \etal proposed DANN to generate task-specific discriminative while domain-wise indiscriminative features \cite{ganin2016domain}. Tzeng \etal presented ADDA for adversarial adaptation \cite{tzeng2017adversarial}.



Both discrepancy and adversarial loss based methods attempt to match the whole source and target domain distribution completely, neither of them considers the target domain data structure and task-specific decision boundaries. To address this, Saito \etal adopted the task-specific category decision boundaries and proposes a model with two classifiers as a discriminator to detect the relationship between the source and target domain data (MCD) \cite{saito2018maximum}. By maximizing the prediction results of the two classifiers, the framework is able to screen out target samples that are near the category decision boundaries and far from the source domain support. Following this, 
Lee \etal extended MCD and proposed a novel Wasserstein metric to capture the natural notion of dissimilarity between the outputs of two task-specific classifiers \cite{lee2019sliced}. Most recently, Li \etal claimed that label distribution alignment is still not enough and present Joint Adversarial Domain Adaptation (JADA) to simultaneously align domain-wise and class-wise distributions across source and target in a unified adversarial learning process \cite{li2019joint}. Unfortunately, existing works seek to maximize the prediction difference between two same architecture classifiers to explore different task-specific knowledge, limiting the divergence of category decision boundaries captured across domains.

Differently, we propose a novel framework with two different structure classifiers, which assist the model to learn more diverse data distribution patterns and less similar category decision boundaries from different perspectives. Integrating task-specific category boundaries and feature-level cross-domain adaptation, our proposed model is able to narrow the data mismatch of source and target domain in the shared domain invariant embedding space.  Moreover, we explore a cross-domain discriminative distribution alignment under the sample Importance Guided Optimization strategy, which has been experimentally proven to eliminate the source and target domain shift.

\section{The Proposed Method}


\subsection{Preliminaries and Motivation}

Given a labeled source domain $\mathcal{D}_s = \{ \mathbf{X}_s, \mathbf{Y}_s \} = \{ (\mathbf{x}_s^i, \mathbf{y}_s^i)\}_{i=1}^{n_s}$ which contains $n_s$ labeled samples, as well as an unlabeled target domain $\mathcal{D}_t = \{ \mathbf{X}_t\} = \{\mathbf{x}_t^j \}_{j=1}^{n_t}$ of $n_t$ unlabeled samples. 
$\mathbf{P}_s(\mathbf{x}_s)$ and $\mathbf{P}_t(\mathbf{x}_t)$ denote the source and target domain different data distributions respectively ($\mathbf{P}_s(\mathbf{x}_s) \neq \mathbf{P}_t(\mathbf{x}_t)$). $\mathcal{C}_s$ and $\mathcal{C}_t$ mean source and target domain identical label spaces. $\mathbf{Y}_s \in \mathbb{R}^{n_s \times C}$ is the source domain ground truth label set which is accessible for training, where $C = C_{s/t} = |\mathcal{C}_{s/t}|$ is the number of total categories. The goal of domain adaption is to seek a model to predict the unlabeled target data over the supervision from the source domain.

Recent domain adaptation works apply adversarial networks to generate domain invariant features of the source and target domain samples, which will make the classifiers trained only on the source domain data available on the target domain\cite{ganin2014unsupervised,ghifary2016deep,zhang2018collaborative}. Most of them aim to match the distribution of source and target domain completely, without considering the task-specific decision boundaries between different categories. Most recently, the idea of dual adversarial classifiers \cite{saito2018maximum,lee2019sliced,zhang2019domain,li2019joint} has been explored to replace the original adversarial domain adaptation with a binary domain discriminator. However, they obtain two same-type classifiers from scratch over labeled source data. This would limit the discriminative ability in target prediction since the same-type classifiers would tend to have similar properties. Traditional neural networks classifier aims to fit the training data by achieving optimal objective value, thus the learned classifier boundaries would capture the global structure of the data to maximally separate different classes. Such a decision boundary over source supervision cannot be well adapted to target samples in different distribution. Therefore, two same-architecture neural network classifiers over source supervision are challenging to diversify the decision boundaries.

This motivates us to explore two different architecture classifiers, and thus we propose a novel adversarial dual classifiers network with two different structure classifiers, Neural Networks Classifier $C_N(\cdot)$ and Prototypical Classifier $C_P(\cdot)$ \cite{snell2017prototypical}, which can capture various data distribution pattern and more diverse task-specific category boundaries from different perspectives, and also promote the out of source support target samples detection process.
Interestingly, the prototypical classifier explores the local structure of the data since prototypes are used to assign labels based on the similarity between samples and each prototype. The competition between two different structure classifiers is more likely to diversify the decision boundaries to benefit from adversarial training with domain-invariant generator.

\subsection{Adversarial Dual Distinct Classifiers Network}


We first present the overall framework of our proposed adversarial dual classifier network in Fig. \ref{fig:framework}. Given the labeled source and unlabeled target domain data, the domain invariant embedding features are generated and aligned by the discriminative cross-domain alignment, then the dual classifiers, which consist of two classifiers with different architectures, will promote the task-specific decision boundaries further. $G(\cdot)$ is the generator used to map source and target domain data to a shared embedding feature space, in which the target samples are close to the support of the source domain data. The following two different structure classifiers, fully-connected neural network classifier $C_N(\cdot)$ and prototypical classifier $C_P(\cdot)$, will capture diverse and various task-specific categories knowledge on target domain from different perspectives.


\subsubsection{Dual Classifiers Over Source Supervision}
\label{dualclassifier}

Since $\mathbf{X}_s$ and $\mathbf{X}_t$ have different distributions, a domain-invariant feature generator $G(\cdot)$ is deployed to capture more enriched information across source and target through hierarchical structures, followed by our dual classifiers, $C_N(\cdot)$ (fully-connected neural network classifier) and $C_P(\cdot)$ (prototypical classifier). With the extracted feature $ \mathbf{z}_{s/t}^i = G(\mathbf{x}_{s/t}^i)$ from $G(\cdot)$ as input, we can calculate the corresponding probability prediction with two classifiers $C_N(\cdot)$ and $C_P(\cdot)$ as $\mathbf{\hat{y}}_{N/P,s/t}^i=C_{N/P}(\mathbf{z}_{s/t}^i)$.

Specifically, $C_N(\cdot)$ is the traditional multi-layer non-linear classifier, while $C_P(\cdot)$ is defined as the similarity between target sample feature $\mathbf{z}_t^i$ to each category prototype $\boldsymbol{\mu}_c$ (i.e., class center), that is, $\mathbf{\hat{y}}_{P,t}^{i(c)}=\mathbf{\Phi}\big(\mathbf{z}_t^i,\boldsymbol{\mu}_c\big)$.
For each class, the prototype $\boldsymbol{\mu}_c = \frac{1}{n_t^c} \sum_{i=1}^{n_t^c} \mathbf{z}_t^{i(c)}$, where $n_t^c$ and $\mathbf{z}_t^{i(c)}$ denote the number of target samples and extracted domain invariant feature belonging to class $c$. We apply the $C_P(\cdot)$ prediction $\mathbf{\hat{y}}_{P,t}^i$ as the predicted pseudo label to target sample $\mathbf{x}_t^i$ to get the category prototypes $\boldsymbol{\mu}_c$.


In order to obtain task-specific discriminative features from generator $G(\cdot)$, while keeping classification performance on source domain, we add the supervision from source to learn the parameters of $C_N(\cdot)$ and $G(\cdot)$. Since $C_P(\cdot)$ does not contain any trainable parameters, the supervision over $C_P(\cdot)$ prediction on the source domain tends to optimize the generator $G(\cdot)$ only. To this end, we aim to minimize the cross-entropy loss over $\mathbf{Y}_s$ and predicted labels from $C_N(\cdot)$ and $C_P(\cdot)$, defined as follows:
\begin{equation}
\begin{aligned}
    \label{loss_s}
    \mathcal{L}_s = \frac{1}{n_s}\sum_{i=1}^{n_s} \mathcal{L}(\mathbf{\hat{y}}_{N,s}^{i}, \mathbf{y}_s^i)+\frac{1}{n_s}\sum_{i=1}^{n_s} \mathcal{L}(\mathbf{\hat{y}}_{P,s}^{i}, \mathbf{y}_s^i),
\end{aligned}
\end{equation}
where $\mathcal{L}$ is the cross-entropy loss. $\mathbf{\hat{y}}_{N,s}^{i}$ and $\mathbf{\hat{y}}_{P,s}^{i}$ are the probability outputs of classifier $C_N(\cdot)$ and $C_P(\cdot)$, while $\mathbf{y}_s^i$ is the ground-truth label of source sample $\mathbf{x}_s^i$, respectively.

\subsubsection{Adversarial Dual Classifiers}


The dual classifiers are capable of recognizing target domain samples close to the support of the source domain. For those target domain samples which are far from the source domain support, the two classifiers would tend to obtain different probability outputs. To detect target samples outside of the support from source supervision, we propose to measure the disagreement of the classifiers prediction results with distribution discrepancy measurement \cite{lee2019sliced,saito2018maximum}. 

Existing works exploit varying the dual classifiers by maximizing the divergence between the predictions. However, the same classifier structure with slightly different random initializations \cite{saito2018maximum,lee2019sliced} will weaken the ability to capture diverse task-specific knowledge and decision boundaries from different perspectives.
In our model, we build two different architecture classifiers, which are more likely to capture the inconsistent information from various perspective. Thus, adversarial training would further enhance the target prediction performance, and the classifier discrepancy is defined as:
\begin{equation}
\label{eq:classifier-discrepancy}
\mathcal{L}_{dis} = \mathcal{F}(\mathbf{\hat{y}}^i_{N,t},\mathbf{\hat{y}}^i_{P,t}),
\end{equation}
where 
$\mathbf{\hat{y}}^i_{N/P,t}$ represent the probability prediction obtained from the two classifiers for the sample $\mathbf{x}^i_t$ respectively. $\mathcal{F}(\cdot,\cdot)$ denotes the discrepancy measurement function, 
which is able to capture distribution geometric information to calculate the discrepancy between the probability prediction distributions, and solve gradient vanishing problems occurred in adversarial learning methods.

\subsubsection{Discriminative Cross-Domain Alignment}
\label{cmmd}

So far, our model only aligns cross-domain distributions in terms of label space, we further exploit feature distribution alignment to boost the domain-invariant feature learning. 
Empirical Maximum Mean Discrepancy (MMD) has been verified as a promising technique to minimize the domain-wise mean of two domains or class-wise mean with the pseudo labels of the target \cite{long2013transfer}. The domain-wise MMD to measure marginal distribution across the source and target domains is defined as $\mathcal{H}(\mathbb{E}_{\mathbf{x}_s^i  \sim \mathcal{D}_s}[\mathbf{z}_s^i]-\mathbb{E}_{\mathbf{x}_t^j  \sim \mathcal{D}_t}[\mathbf{z}_t^j])$ \cite{long2013transfer}, where $\mathcal{H}(\cdot)$ is the function used to evaluate the distribution difference. Furthermore, existing works \cite{ding2018graph} also seek to explore the class-wise MMD to align conditional distribution disparity across domain:
\begin{equation}
\label{cmmd}
\begin{aligned}
\mathcal{L}_c = \frac{1}{C} \sum_{c=1}^{C}\mathcal{H}\Big(\mathbb{E}_{\mathbf{x}_s^i  \sim \mathcal{D}_s^c}[\mathbf{z}_s^i]-\mathbb{E}_{\mathbf{x}_t^j  \sim \mathcal{D}_t^c}[\mathbf{z}_t^j]\Big),
\end{aligned}
\end{equation}
where $C$ denotes the total number of categories, $\mathbf{z}_{s/t}^{i/j}$ denote the generated embedding representations of source sample $\mathbf{x}_{s}^{i}$ and target sample $\mathbf{x}_{t}^{j}$ belonging to class $c$. 

However, conventional DA algorithms only seek to minimize the distribution difference between source and target domains when samples are from the same class. We further propose to explicitly take the information of different categories into account and measure the \textit{diff-class} divergence across domains defined as:
\begin{equation}
\label{cmmd}
\begin{aligned}
\mathcal{L}_{d} =  \dfrac{1}{C}\dfrac{1}{C-1} \sum\limits_{c=1}^{C} \sum\limits_{\substack{c'=1, \\c' \neq c}}^{C} \mathcal{H}\Big(\mathbb{E}_{\mathbf{x}_s^i  \sim \mathcal{D}_s^c}[\mathbf{z}_s^i]-\mathbb{E}_{{\mathbf{x}_t^j}  \sim \mathcal{D}_t^{c'}}[\mathbf{z}_t^j]\Big),
\end{aligned}
\end{equation}
where the \textit{diff-class} divergence $\mathcal{L}_{d}$ calculates the average distances of all different class center pairs across domains. To sum up, our discriminative cross-domain alignment is defined as $\mathcal{L}_{m} = \mathcal{L}_{c} - \mathcal{L}_{d}$. 

Due to the lack of target domain labels, we explicitly assign $\mathbf{\hat{y}}_{P,t}^i$, the prediction of $C_P(\cdot)$, as pseudo labels to the target samples $\mathbf{x}_t^i$. To exploit more effective knowledge transfer iteratively, we propose an \textit{Importance Guided Optimization} strategy to only consider those target samples with high prediction confidences during the cross-domain alignment since lower-confident samples would mislead the optimization. That is, only samples with $\{(\mathbf{x}_t^i, \mathbf{\hat{y}}_{P,t}^{i(c)}) \mid \hat{y}_{P,t}^{i(c)} > \sigma_1, \mathbf{x}_t^i \in \mathcal{D}_t \}$ are accepted to construct the cross-domain alignment $\mathcal{L}_m$, where $\mathbf{\hat{y}}_{P,t}^{i(c)}$ is the $C_P(\cdot)$ probability prediction of $\mathbf{x}_t^i$ belonging to class $c$, and $\sigma \in [0,1]$ is a constant threshold. It is noteworthy that we do not impose always covering the whole label space during training, since only considering those classes with high-confident samples is prone to result in effective cross-domain alignment by avoiding too many mis-classified target samples, especially in the early training stage.



\subsection{Overall Objective and Optimization}

To eliminate the side effect of uncertainty on unlabeled target prediction, we also explore the entropy minimization regularization \cite{zhang2019domain,long2018conditional,long2016unsupervised}:
\begin{equation}
    \mathcal{L}_{em} = -\frac{1}{n_t} \sum_{i=1}^{n_t}\sum_{c=1}^{C} (\mathbf{\hat{y}}_{N,c}^{i} \log \mathbf{\hat{y}}_{N,c}^{i} + \mathbf{\hat{y}}_{P,c}^{i} \log \mathbf{\hat{y}}_{P,c}^{i}),
\end{equation}
where $\mathbf{\hat{y}}_{N,c}^{i}$ and $\mathbf{\hat{y}}_{P,c}^{i}$ 
denote the prediction of $\mathbf{x}_t^i$ belonging to class $c$ obtained by $C_N(\cdot)$ and $C_P(\cdot)$, respectively. 

To sum up, we integrate adversarial dual classifiers training and cross-domain discriminative alignment together, and propose our overall objective function as:
\renewcommand{\arraystretch}{1.2}
\begin{equation}
\begin{array}{c}
    \min \limits_{G} \mathcal{L}_s + \mathcal{L}_{em} + \lambda_1  \mathcal{L}_{dis} + \lambda_2 \mathcal{L}_{m}, \\
    \min \limits_{C_N} \mathcal{L}_s  - \lambda_1 \mathcal{L}_{dis},
\end{array}
\end{equation}
where $\lambda_1$ and $\lambda_2$ are hyper-parameters to balance the contribution of loss terms $\mathcal{L}_{dis}$, $\mathcal{L}_{m}$, respectively.

Similar to existing adversarial networks training strategy, we freeze the generator $G(\cdot)$ to train classifiers, then freeze the parameters of the classifiers to update the generator $G(\cdot)$. It is noteworthy that only $C_N(\cdot)$ contains trainable parameters because $C_P(\cdot)$ only relies on the embedding features produced by the generator $G(\cdot)$. Meanwhile, inspired by \cite{saito2018maximum}, in order to keep the performance of the networks on the source domain and detect target samples far from source domain support, we train our framework by three steps:

\vspace{1mm}\noindent\textbf{Step A.} We train the feature generator $G(\cdot)$ and classifier $C_N(\cdot)$ only on source domain $\mathcal{D}_s$ which is the same as supervised learning tasks. Due to $C_P(\cdot)$ does not have any trainable parameters, only parameters in $G(\cdot)$ and $C_N(\cdot)$ would be updated. Our model aims to detect target samples which are outside of source support from those which are close to support of source domain, keeping good ability and performance on classifying the source domain samples correctly is crucial and necessary. The optimization objective is defined as $\min \limits_{G,C_N} \mathcal{L}_s$.

\noindent\textbf{Step B.} We need to assign unlabeled target domain samples pseudo labels by classifiers we already have. In our experiments, we explore the prediction results of $C_P(\cdot)$ to obtain pseudo labels of the target samples, which are experimentally proven to achieve better performance, and we will discuss it in the ablation analysis section. We fix the feature generator $G(\cdot)$ and update the classifier $C_N(\cdot)$ to maximize the distribution discrepancy between the classification results of $C_N(\cdot)$ and $C_P(\cdot)$ on the target domain, which can detect the target samples excluded by the source domain data support, and we obtain the training objective function as $\min \limits_{C_N} \mathcal{L}_s - \lambda_1 \mathcal{L}_{dis}$.

\noindent\textbf{Step C.} We freeze the parameters of the classifier $C_N(\cdot)$ and update generator $G(\cdot)$ to minimize the distribution discrepancy between the predictions of $C_N(\cdot)$ and $C_P(\cdot)$ on the target domain, through which both $C_N(\cdot)$ and $C_P(\cdot)$ classifiers will have more similar and correct prediction on target domain samples. Furthermore, together with the discriminative cross-domain alignment, the generator $G(\cdot)$ 
tends to couple the source and target domain closer but discriminative in the embedding feature space. The optimization objective is $\min \limits_{G} \mathcal{L}_s + \mathcal{L}_{em} + \lambda_1 \mathcal{L}_{dis} + \lambda_2 \mathcal{L}_{m}$.

 These three steps repeat once in each iteration in our experiments. The generator $G(\cdot)$ and classifier $C_N(\cdot)$ are initialized and pre-trained on source domain data. 

\section{Experimental Results}

\begin{table*} [!h]
\linespread{1.15} 
\centering
\caption{Comparisons of Recognition Rates ($\%$) of Unsupervised Domain Adaptation on Office+Home Dataset (ResNet-50).} 
\label{OfficeHome}
\setlength{\tabcolsep}{1pt} 
\renewcommand{\arraystretch}{1} 
\begin{tabular}{c|cccccccccccc|c} 
\toprule
Method & Ar$\rightarrow$Cl & Ar$\rightarrow$Pr & Ar$\rightarrow$Rw & Cl$\rightarrow$Ar & Cl$\rightarrow$Pr & Cl$\rightarrow$Rw & Pr$\rightarrow$Ar & Pr$\rightarrow$Cl & Pr$\rightarrow$Rw & Rw$\rightarrow$Ar & Rw$\rightarrow$Cl & Rw$\rightarrow$Pr & Avg. \\   
\hline
Res-50 \cite{he2016deep}  & 34.9 & 50.0 & 58.0 & 37.4 & 41.9 & 46.2 & 38.5 & 31.2 & 60.4 & 53.9 & 51.2 & 59.9 & 46.1 \\
DAN  \cite{long2015learning}    & 43.6 & 57.0 & 67.9 & 45.8 & 56.5 & 60.4 & 44.0 & 43.6 & 67.7 & 63.1 & 51.5 & 74.3 & 56.3 \\
RevGrad \cite{ganin2014unsupervised} & 45.6 & 59.3 & 70.1 & 47.0 & 58.5 & 60.9 & 46.1 & 43.7 & 68.5 & 63.2 & 51.8 & 76.8 & 57.6 \\
JAN \cite{long2017deep}     & 45.9 & 61.2 & 68.9 & 50.4 & 59.7 & 60.0 & 45.8 & 43.4 & 70.3 & 63.9 & 52.4 & 76.8 & 58.3 \\
SE \cite{french2017self}      & 48.8 & 61.8 & 72.8 & 54.1 & 63.2 & 65.1 & 50.6 & 49.2 & 72.3 & 66.1 & 55.9 & 78.7 & 61.5 \\
DSR \cite{cai2019learning}     & 53.4 & 71.6 & 77.4 & 57.1 & 66.8 & 69.3 & 56.7 & 49.2 & 75.7 & 68.0 & 54.0 & 79.5 & 64.9 \\
DWT-MEC \cite{roy2019unsupervised} & 50.3 & 72.1 & 77.0 & 59.6 & 69.3 & 70.2 & 58.3 & 48.1 & 77.3 & 69.3 & 53.6 & 82.0 & 65.6 \\
CDAN+E \cite{long2018conditional}  & 50.7	& 70.6 & 76.0 & 57.6 & 70.0 & 70.0 & 57.4 & 50.9 & 77.3 & 70.9 & 56.7 & 81.6 & 65.8 \\
MCS  \cite{Liang2019DistantSC}    & \underline{55.9} & \underline{73.8} & \underline{79.0} & 57.5 & 69.9 & 71.3 & 58.4 & 50.3 & 78.2 & 65.9 & 53.2 & 82.2 & 66.3 \\
AFN \cite{xu2019larger}& 52.0 & 71.7 & 76.3 & 64.2 & 69.9 & 71.9 & 63.7 & \underline{51.4} & 77.1 & 70.9 & \underline{57.1} & 81.5 & 67.3 \\
SymNets	\cite{zhang2019domain}&47.7&	72.9&	78.5&	64.2&	71.3 &	\underline{74.2}&	\underline{64.2}&	48.8&	79.5 &	\underline{74.5}&	52.6&	81.6&	67.6\\
BDG \cite{yang2020bi} & 51.5 & 73.4 & 78.7 & \textbf{65.3} & \underline{71.5} & 73.7 & \textbf{65.1} & 49.7 & \textbf{81.1} & \textbf{74.6} & 55.1 & \textbf{84.8} & \underline{68.7} \\
\hline
Ours & \textbf{57.4} & \textbf{77.3} & \textbf{80.0} & \underline{63.4} & \textbf{76.4} & \textbf{76.4}	& \underline{64.2} & \textbf{52.4} & \underline{80.7} & 69.6 & \textbf{57.2} & \underline{83.9} & \textbf{69.9} \\

\bottomrule
\end{tabular} \vspace{-1mm}
\end{table*}

\begin{table*} [t]
\linespread{1.15} 
\centering\small
\caption{Comparisons of Recognition Rates ($\%$) of Unsupervised Domain Adaptation on Office-31 Dataset (ResNet-50).} 
\label{Office-31}
\setlength{\tabcolsep}{2pt} 
\renewcommand{\arraystretch}{1} 
\begin{tabular}{c|ccccccccc|c} 
\toprule
Method & Res-50 \cite{he2016deep} & DAN \cite{long2015learning} & RevGrad \cite{ganin2014unsupervised} & JAN \cite{long2017deep} & MADA \cite{pei2018multi} & CDAN+E \cite{long2018conditional} & AFN \cite{xu2019larger}& SymNets \cite{zhang2019domain} & BDG \cite{yang2020bi} & Ours \\
  \hline
A$\rightarrow$W & 68.4$\pm$0.2 & 80.5$\pm$0.4 & 82.0$\pm$0.4 & 86.0$\pm$0.4 & 90.0$\pm$0.1 & \textbf{94.1}$\pm$0.1  & 90.1$\pm$0.1 & 90.8$\pm$0.1 & \underline{93.6}$\pm$0.4 & \underline{93.6}$\pm$0.3 \\
D$\rightarrow$W & 96.7$\pm$0.1 & 97.1$\pm$0.2 & 96.9$\pm$0.2 & 96.7$\pm$0.3 & 97.4$\pm$0.1 & 98.6$\pm$0.1  & 98.6$\pm$0.2 & 98.8$\pm$0.3 & \textbf{99.0}$\pm$0.1 & \underline{98.9}$\pm$0.2 \\
W$\rightarrow$D & 99.3$\pm$0.1 & 99.6$\pm$0.1 & 99.1$\pm$0.1 & 99.7$\pm$0.1 & 99.6$\pm$0.1 & \textbf{100.0}$\pm$0.0 & \underline{99.8}$\pm$0.0 & \textbf{100.0}$\pm$0.0 & \textbf{100.0}$\pm$0.0 &\underline{99.8}$\pm$0.0 \\
A$\rightarrow$D & 68.9$\pm$0.2 & 78.6$\pm$0.2 & 79.7$\pm$0.4 & 85.1$\pm$0.4 & 87.8$\pm$0.2 & 92.9$\pm$0.2 & 90.7$\pm$0.5 & \underline{93.9}$\pm$0.5 & 93.6$\pm$0.3 & \textbf{95.4}$\pm$0.3\\
D$\rightarrow$A & 62.5$\pm$0.3 & 63.6$\pm$0.3 & 68.2$\pm$0.4 & 69.2$\pm$0.3 & 70.3$\pm$0.3 & 71.0$\pm$0.3 & 73.0$\pm$0.2 & \underline{74.6}$\pm$0.6 & 73.2$\pm$0.2 &  \textbf{74.9}$\pm$0.3 \\
W$\rightarrow$A & 60.7$\pm$0.3 & 62.8$\pm$0.2 & 67.4$\pm$0.5 & 70.70.5 & 66.4$\pm$0.3 & 69.3$\pm$0.3 & 70.2$\pm$0.3 & \underline{72.5}$\pm$0.5 & 72.0$\pm$0.1 & \textbf{75.0}$\pm$0.5 \\
\hline
Avg. & 76.1 & 80.4 & 82.2 & 84.6 & 85.2 & 87.7  & 87.1 & 88.4 & \underline{88.5} &\textbf{89.6} \\
\bottomrule
\end{tabular} \vspace{-3mm}
\end{table*}

\subsection{Datasets \& Experimental Setup}

\noindent\textbf{Office-Home} \cite{venkateswara2017deep} consists of 15,500 images from 65 categories in 4 different domains: Artistic images (Ar), Clip Art (Cl), Product (Pr), and Real-World images (Rw). In total, by choosing any two domain as one task, we can build 12 cross-domain tasks to evaluate our proposed model.


\vspace{1mm}\noindent \textbf{Office-31} contains 4,110 images of 3 domains: Amazon (A), Webcam (W), and DSLR (D) and each domain consists of 31 categories. We evaluate our method on 6 cross-domain tasks to testify the validation of our model. 


\vspace{1mm}\noindent\textbf{Comparisons.} We compare our proposed method with several state-of-the-art unsupervised domain adaptation models: Deep Adaptation Networks (DAN) \cite{long2015learning}, Reverse Gradient (RevGrad) \cite{ganin2014unsupervised}, Joint Adaptation Networks (JAN) \cite{long2017deep}, Self-Ensembling (SE) \cite{french2017self}, Multi-adversarial Domain Adaptation (MADA) \cite{pei2018multi}, Conditional Adversarial Domain Adaptation Networks (CDAN) \cite{long2018conditional}, Disentangled Semantic Representation (DSR) \cite{cai2019learning}, Domain-specific Whitening Transform \& Min-Entropy Consensus (DWT-MEC) \cite{roy2019unsupervised}, Minimum Centroid Shift (MCS) \cite{Liang2019DistantSC}, Adaptive Feature Norm Approach (AFN) \cite{xu2019larger}, Domain Symmetric Networks (SymNets) \cite{zhang2019domain}, Bi-Directional Generation (BDG) \cite{yang2020bi}. All our experiments follow standard unsupervised domain adaptation protocols: all labeled source domain data and labels, as well as unlabeled target domain data are used for training. All comparisons are back-boned with ResNet-50 or using ResNet-50 features \cite{he2016deep}. 

\vspace{1mm}\noindent\textbf{Implementation Details.} We implement our model with PyTorch and adopt ResNet-50\cite{he2016deep} as the backbone. Specifically, a ResNet-50 network is pre-trained on ImageNet \cite{deng2009imagenet} and fine-tuned on the source domain, then applied to both source and target domain data to obtain the feature representation with dimension 2,048 without the last fully connected layer. $G(\cdot)$ is a two-layer fully-connected neural network, with hidden layer output as 1,024 followed by ReLU activation function, and the dropout probability retaining is 0.5. The output embedding features $\mathbf{z}_{s/t}$ dimension is 512. $C_N(\cdot)$ is a two-layer fully-connected neural network with 512 as the input and hidden layer dimension, the output dimension is the same as the number of categories in the whole label space $C$. Cosine similarity is accepted as the measurement metric $\mathbf{\Phi}(\cdot,\cdot)$ in $C_P(\cdot)$.
All parameters are updated with Adam optimizer \cite{kingma2014adam} and the learning rate is set as 0.001 on Office-Home and Office-31 dataset. $G(\cdot)$ and $C_N(\cdot)$ are pre-trained and initialized on source domain data only with the learning rate as 0.1 for 2,000 iterations. We deploy SWD distance \cite{lee2019sliced} as the discrepancy measurement function $\mathcal{F}(\cdot,\cdot)$, and accept $L$-2 norm as $\mathcal{H}(\cdot)$ to evaluate the distribution divergence. $\lambda_1$ and $\lambda_2$ are fixed as 0.1 for all tasks. $\sigma$ is set to be 0.03. For the prototypical classifier $C_P(\cdot)$, we initialize the class prototypes with the source domain features class centers $\boldsymbol{\mu}_c^s=\frac{1}{n_s^c}\sum_{i=1}^{n_s^c}\mathbf{z}_s^i$, then update the prototypes with target domain category centroids representation $\boldsymbol{\mu}_c^t = \frac{1}{n_t^c}\sum_{j=1}^{n_t^c} \mathbf{z}_t^j$ after obtaining the target domain samples pseudo labels $\mathbf{\hat{y}}_{P,t}$ iteratively till reaching convergence or the max step (which is set as 3), and return the last step $C_P(\cdot)$ prediction.
All results reported in Tables \ref{OfficeHome} and \ref{Office-31} are the average of three random experimental results obtained by classifier $C_P(\cdot)$, and we will discuss the performances of $C_N(\cdot)$ and $C_P(\cdot)$ in the ablation study section.


\begin{figure*}[t]
    \centering
    \includegraphics[width=1\linewidth]{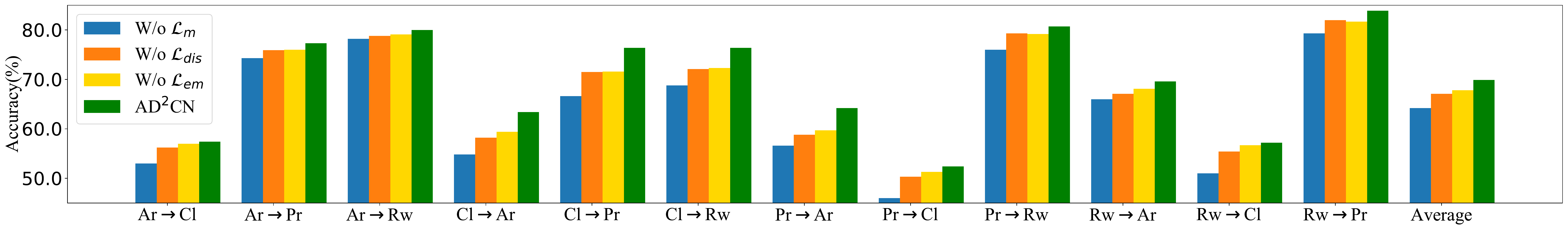}\vspace{-2mm}
    \caption{Ablation experiments about various loss terms contribution  on Office+Home Dataset (ResNet-50).}
    \label{fig:lambda} \vspace{-2mm}
\end{figure*}

\begin{figure*}[t]
    \centering
    \includegraphics[width=1\linewidth]{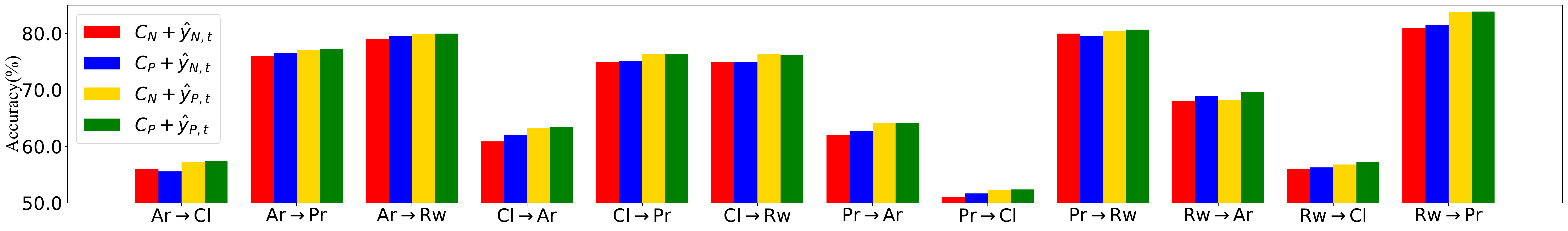}
    \caption{Accuracies of $C_N$ and $C_P$ on Office+Home. red and blue results are obtained with $\mathbf{\hat{y}}_{N,t}$ as target pseudo labels for $\mathcal{L}_m$, the others are based on $\mathbf{\hat{y}}_{P,t}$ as pseudo labels.}
    \label{fig:cncp} \vspace{-4mm}
\end{figure*}

\subsection{Comparison Results}

Table \ref{OfficeHome} and Table \ref{Office-31} report the classification results on target domain data of our proposed model and other comparative methods on Office-Home and Office-31 datasets respectively. All comparison results are from their original paper or quoted from \cite{kurmi2019attending,zhang2019domain,yang2020bi}, as we adopt exactly the same settings. It is noteworthy that our proposed model outperforms state-of-the-art methods on all benchmark datasets in terms of average accuracy, and obtains the best or comparable performances to the state-of-the-art domain adaptation methods in most cases. Although the Office-Home dataset is more challenging than Office-31 due to more categories and samples, as well as significant distribution dissimilarity, our proposed model still improves the performance on most tasks, which demonstrates the efficiency and effectiveness of our proposed framework. 

DAN and JAN are both MMD-based methods, which seek to eliminate the cross-domain distribution disparity and match the whole source and target domain to a shared domain-invariant feature space. DAN attempts to align feature representations from multiple layers through a multi-kernel variant of MMD. JAN aims to transfer joint distributions of multi-layers' activation of the networks across domains. With the help of additional domain adaptation terms (e.g., MMD), DAN and JAN lead to a significant performance boost over the source-only-trained model (i.e., ResNet-50) on most adaptation tasks.


\begin{table} [t]
\linespread{1.14} 
\centering
\caption{$C_N$ v.s. $C_P$ accuracies ($\%$) on Office+Home Ar $\rightarrow$ Cl}
\label{cncp}
\setlength{\tabcolsep}{1pt} 
\renewcommand{\arraystretch}{1} 
\begin{tabular}{c|ccc|ccccc} 
\toprule
 & \multicolumn{3}{c|}{\textit{Balanced}}&\multicolumn{5}{c}{\textit{Imbalanced}} \\
\hline
Y     & Clock & Helmet  &   Knives    &   Bed  &  Couch  &  Folder   &  Marker  &  Pen\\ 
\hline
$n_s$ & 74 & 79 & 72 & 39 & 40 & 20 & 20 & 20  \\ 
$n_t$ & 60 & 69 & 53 & 98 & 64 & 99 & 71 & 99 \\ 
\hline
$C_N$ & \textbf{75.0} & \textbf{71.0} & \textbf{52.8} & 53.1& 67.2 & 25.3 & 18.3 & 51.5 \\ 
$C_P$ & 73.3 & 69.6 & 49.1 & \textbf{55.1} & \textbf{68.8} & \textbf{28.3} & \textbf{21.1} & \textbf{53.5} \\ 

\bottomrule
\end{tabular}\vspace{-3mm}
\end{table}

\begin{table} [t]
\linespread{1.15} 
\centering\small
\caption{Comparisons of Dual Classifiers Structure Influence to Recognition Rates ($\%$) of Unsupervised Domain Adaptation on Office-31 Dataset (ResNet-50).} 
\label{mcd}
\setlength{\tabcolsep}{1.5pt} 
\renewcommand{\arraystretch}{1} 
\begin{tabular}{cccccccc} 
\toprule
  Method & A$\rightarrow$W & D$\rightarrow$W & W$\rightarrow$D & A$\rightarrow$D & D$\rightarrow$A & W$\rightarrow$A & Avg. \\
  \hline
MCD \cite{saito2018maximum} & 88.6 & 98.5 & 100.0 &92.2 & 69.5& 69.7& 86.5\\
SWD \cite{lee2019sliced} & 90.4& 98.7 & 100.0& 94.7 & 70.3 & 70.5 & 87.4 \\
\hline

Ours (same)  & 93.3 & 98.8& \textbf{100}& 94.7 & 72.4 & 73.6 & 88.8\\
Ours & \textbf{93.6}  &  \textbf{98.9} & 99.8 & \textbf{95.4} & \textbf{74.9} & \textbf{75.0} & \textbf{89.6}\\
\bottomrule
\end{tabular}\vspace{-5mm}
\end{table}







RevGrad implements adversarial networks and applies gradient reversal layer to train a domain discriminator. CDAN and MADA both exploit multiplicative interactions between feature representations and category predictions as high-order features to promote the adversarial training. SE explores the use of self-ensembling for visual domain adaptation. DSR assumes that the data generation process is controlled by the semantic latent variables and domain latent variables independently, so employs a variational auto-encoder in order to reconstruct them. MCS designs a unified framework without accessing the source domain data and iteratively assigns pseudo labels to the target samples by an alternating minimization scheme. 

DWT-MEC proposes domain alignment layers with feature whitening to match source and target domain distributions and leverages the unlabeled target data by Min-Entropy Consensus loss. AFN proposes a novel Adaptive Feature Norm approach to progressively adapting the feature norms of the two domains to a large range of values.
SymNets exploits a novel adversarial classifiers networks and a two-level domain confusion scheme driving the learning of categories invariant intermediate features across domains. BDG bridges source and target domain through consistent classifiers interpolating two intermediate domains.


\subsection{Ablation Analysis}
\label{ablation}
In this section, we analyze the contribution and influence of several important terms and hyper-parameters sensitivity in our proposed model.

\begin{figure*}[t]
    \centering
    \includegraphics[width=1\linewidth]{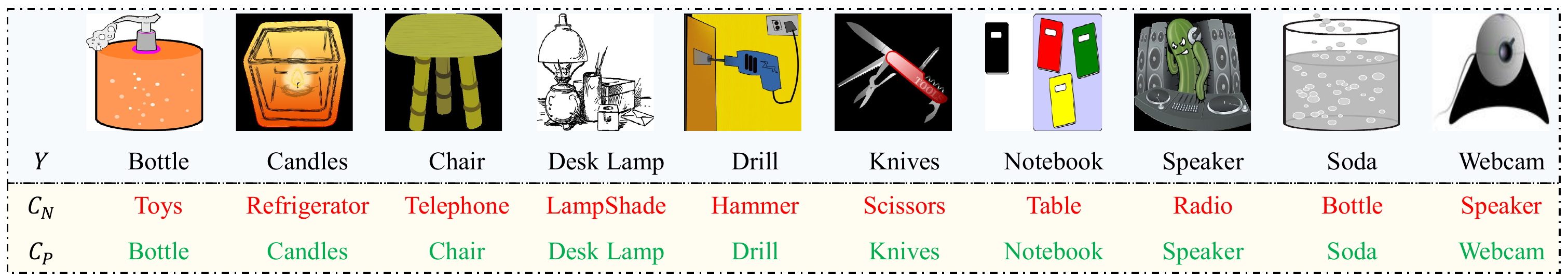}
    \caption{Ten Samples from Office-Home Ar$\rightarrow$Cl. $Y$ row denotes the ground-truth labels, $C_N$ row shows the mis-classified labels, while $C_P$ means the correctly prediction.}
    \label{fig:samples_noborder}\vspace{-3mm}
\end{figure*}

\begin{figure*}[t]
    \centering
    \includegraphics[width=1\linewidth]{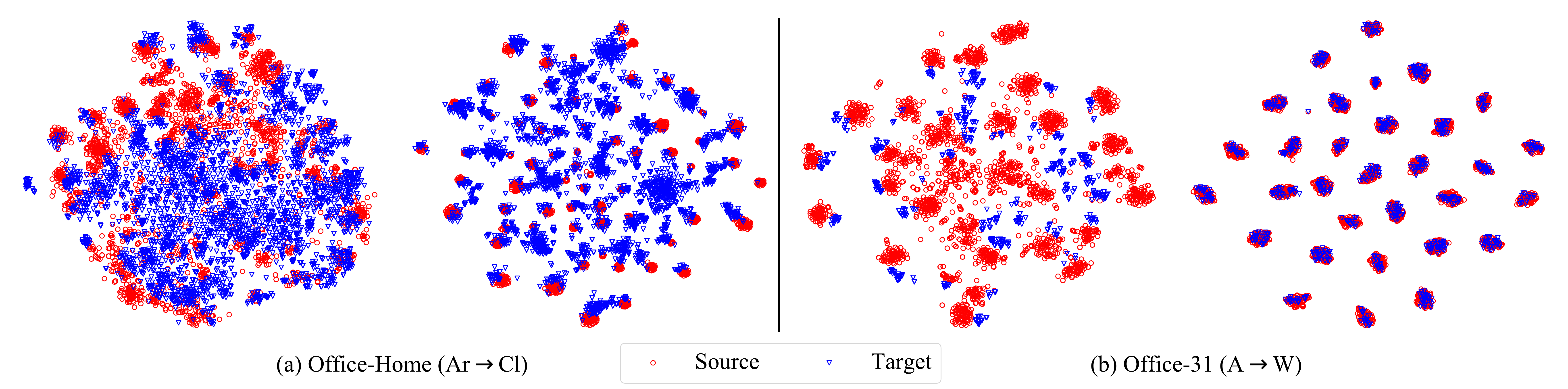}\vspace{-1mm}
    \caption{t-SNE visualization of source and target samples features before (left column) and after (right column) domain adaptation through our proposed model. (a) shows the task of Ar$\rightarrow$Cl from Office-Home and (b) reports the task of A$\rightarrow$W from Office-31.}
    \label{fig:tSNE}\vspace{-4mm}
\end{figure*}



\begin{figure}[t]
    \centering
    \includegraphics[width=1\linewidth]{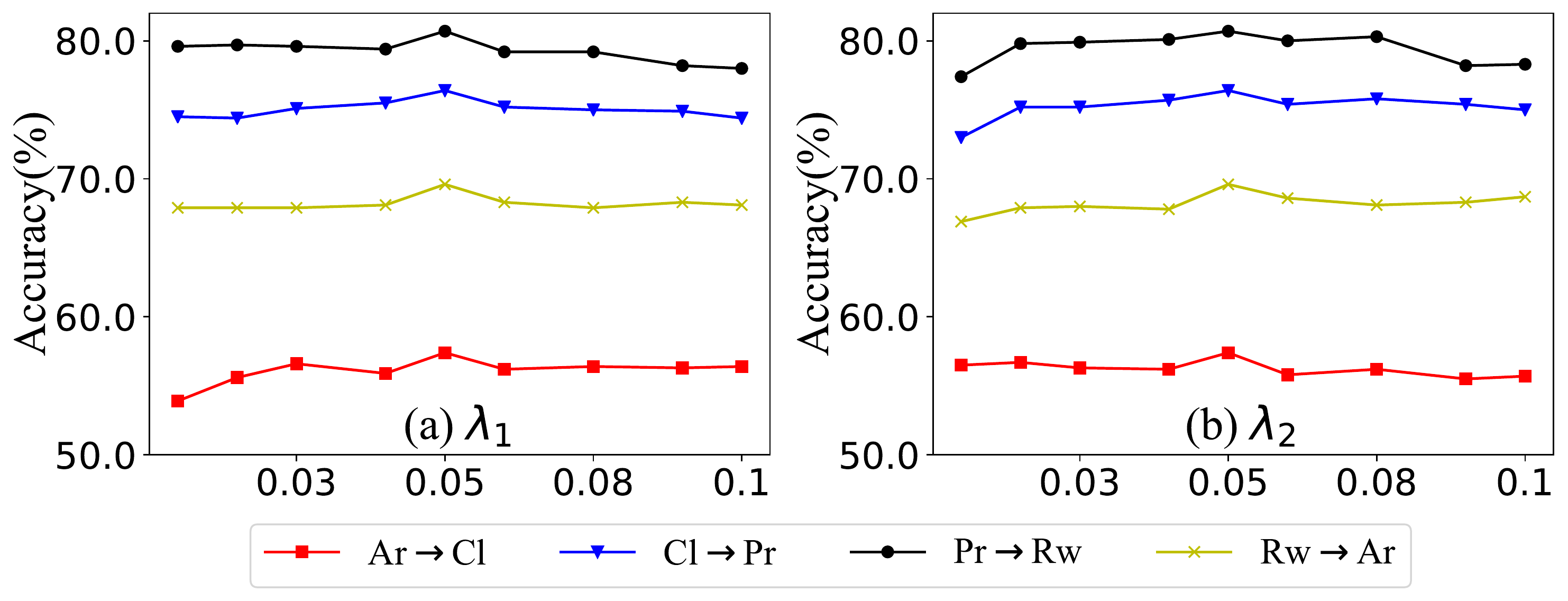}
    \caption{Parameters sensitivity analysis on 4 different tasks from Office-Home dataset of (a) $\lambda_1$ and (b) $\lambda_2$ }
    \label{fig:params}\vspace{-5mm}
\end{figure}

First, we discuss the influence of each component in our framework. By removing one of $\mathcal{L}_{dis}$, $\mathcal{L}_{m}$, and $\mathcal{L}_{em}$, while keeping other terms same as original AD$^2$CN, we obtain three variants AD$^2$CN w/o $\mathcal{L}_{dis}$, 
 AD$^2$CN w/o $\mathcal{L}_{m}$, and AD$^2$CN w/o $\mathcal{L}_{em}$.
From Fig. \ref{fig:lambda}, we notice that all three components contribute to improving the domain adaptation performance, while our proposed discriminative cross-domain alignment $\mathcal{L}_{m}$ plays a more crucial role than others, i.e., discrepancy and entropy minimization loss.

Secondly, we compare the performances of $C_N(\cdot)$ and $C_P(\cdot)$ while accepting $\mathbf{\hat{y}}_{N,t}$ or $\mathbf{\hat{y}}_{P,t}$ as target domain pseudo labels for $\mathcal{L}_m$. From the results in Fig. \ref{fig:cncp}, we observe that results with $\mathbf{\hat{y}}_{P,t}$ as pseudo labels are better than the results with $\mathbf{\hat{y}}_{N,t}$ in most cases. Compared to $C_N(\cdot)$, which is trained on the source domain, $C_P(\cdot)$ is based on the target prototypes and keeps better performance even on the early training stage. Fig. \ref{fig:samples_noborder} shows several test samples that $C_P(\cdot)$ classifies correctly while $C_N(\cdot)$ cannot handle, which emphasizes the superiority of $C_P(\cdot)$.

Thirdly, we discuss the necessity and effectiveness of two different types of classifiers in our framework. Table \ref{cncp} shows the selective target domain class-wise recognition accuracy on OfficeHome Ar $\rightarrow$ Cl case produced by the two classifiers $C_N$ and $C_P$ in our proposed model, as well as the number of samples in each class from the source and target domains. From the results we notice that for the categories having sufficient well labeled source samples as well as balanced target domain samples for training, $C_N$ have better performance than $C_P$, while for other categories with imbalanced distribution across domains and insufficient labeled source samples for training, $C_P$ always performs better than $C_N$. The observation proves that for imbalanced dataset, $C_N$ and $C_P$ have different speciality for different categories with various cross-domain distributions. More over, we show the comparison results of MCD \cite{saito2018maximum}, SWD\cite{lee2019sliced}, and our proposed model on Office-31 dataset in Table \ref{mcd}. MCD and SWD are two dual classifier adversarial frameworks for domain adaptation, but using two completely same structure neural networks classifiers. We also replace the $C_N$ and $C_P$ in our proposed model with two same structure neural networks classifiers and report the results as Ours(same). It is noteworthy that our proposed model achieves the best performance on most cases as well as the average accuracy compared to other same classifier structure methods, which proves the effectiveness and necessity of applying two distinct architecture classifiers.

Fourthly, we visualize the t-SNE embeddings (Fig. \ref{fig:tSNE}) of feature representations generated by $G(\cdot)$ before and after the domain adaptation through our proposed model, in which each category is represented as a cluster and different colors denote the different domains. Before adaptation, the source and target domains are totally mismatched, while our method shows the promising ability to make inter-class separated and intra-class clustered tightly.


Finally, we analyze the sensitivity of $\lambda_1$ (Fig. \ref{fig:params} (a))and $\lambda_2$ (Fig. \ref{fig:params} (b)) by listing four tasks from Office-Home dataset (Ar $\rightarrow$ Cl, Cl $\rightarrow$ Pr, Pr $\rightarrow$ Rw, Rw $\rightarrow$ Ar). Specifically, we set the ranges of $\lambda_1$ and $\lambda_2$ from 0.001 to 0.2, and evaluate one by fixing the other one as 0.1. From the results, we notice the accuracy curves are almost flat and stable, which indicates our proposed model is not sensitive to the values of $\lambda_1$ nor $\lambda_2$.


\section{Conclusion}

We presented a novel Adversarial Dual Distinct Classifier Networks (AD$^2$CN) for unsupervised domain adaptation to align source and target domain distribution discrepancy as well as task-specific category boundaries. Specifically, we designed two different architecture classifiers to detect target samples excluded by the source domain support by aligning the task-specific decision boundaries obtained by the two classifiers. Meanwhile, a domain-invariant feature generator was proposed to embed source and target domain data to a shared feature space under the guidance of discriminative cross-domain alignment. We evaluated our proposed model on two cross-domain visual benchmarks and obtained better performance over state-of-the-art methods, proving the effectiveness of our method.

\clearpage
\balance
{\small
\bibliographystyle{ieee_fullname}
\bibliography{main}
}

\end{document}